\documentclass[A4paper, 11 pt, conference]{ieeeconf}  
\IEEEoverridecommandlockouts       
\usepackage{geometry}
\usepackage{mars}
\usepackage{graphics} 
\usepackage{epsfig} 
\usepackage{mathptmx} 
\usepackage{times} 
\usepackage{amsmath} 
\usepackage{amssymb}  
\usepackage{subfigure}
\usepackage{algorithm} 
\usepackage{algorithmic} 
\usepackage{multirow} 
\usepackage{color}
\usepackage{hyperref}
\usepackage{tabularx}
\usepackage{float}
\usepackage{booktabs}
\begin{document}

\title{\LARGE \bf
Empowering Robot Path Planning with Large Language Models: osmAG Map Topology \&  Hierarchy Comprehension with LLMs}

\author{Fujing Xie$^*$, S\"{o}ren Schwertfeger$^*$$^\dagger$
\thanks{
	$^*$The authors are with the Key Laboratory of Intelligent Perception and Human-Machine Collaboration -- ShanghaiTech University, Ministry of Education, China.{ \{xiefj,soerensch\}@shanghaitech.edu.cn}
}
\thanks{
	$^\dagger$
	S\"{o}ren Schwertfeger is the corresponding author.
}
\thanks{
	This work has been partially funded by the Shanghai Frontiers Science Center of Human-centered Artificial Intelligence.
	This work was also supported by the Science and Technology Commission of Shanghai Municipality (STCSM), project 22JC1410700 "Evaluation of real-time localization and mapping algorithms for intelligent robots". The experiments of this work were supported by the core facility Platform of Computer Science and Communication, SIST, ShanghaiTech University.
}
}

%


\marsPublishedIn{Accepted for:} 		

\marsVenue{IEEE International Conference on Robotics and Biomimetics (ROBIO)}

\marsYear{2024}

\marsPlainAutors{Fujing Xie, S\"oren Schwertfeger}


\marsMakeCitation{Empowering Robot Path Planning with Large Language Models: osmAG Map Topology Hierarchy Comprehension with LLMs}{IEEE Press}


\marsIEEE{}


\makeMARStitle

%
\maketitle 
\thispagestyle{empty}

\begin{abstract}
Large Language Models (LLMs) have demonstrated great potential in robotic applications by providing essential general knowledge. Mobile robots rely on map comprehension for tasks like localization and navigation. In this paper, we explore enabling LLMs to comprehend the topology and hierarchy of Area Graph, a text-based hierarchical, topometric semantic map representation utilizing polygons to demark areas such as rooms or buildings.
Our experiments demonstrate that with the right map representation, LLMs can effectively comprehend Area Graph's topology and hierarchy. After straightforward fine-tuning, the LLaMA2 models exceeded ChatGPT-3.5 in mastering these aspects.
 Our dataset, dataset generation code, fine-tuned LoRA adapters can be accessed at
\href{https://github.com/xiefujing/LLM-osmAG-Comprehension}{https://github.com/xiefujing/LLM-osmAG-Comprehension}.
\end{abstract}
\vspace{-3mm}
\section{Introduction}
\begin{figure}[t]
	\centering
	\includegraphics[width=0.43\textwidth]{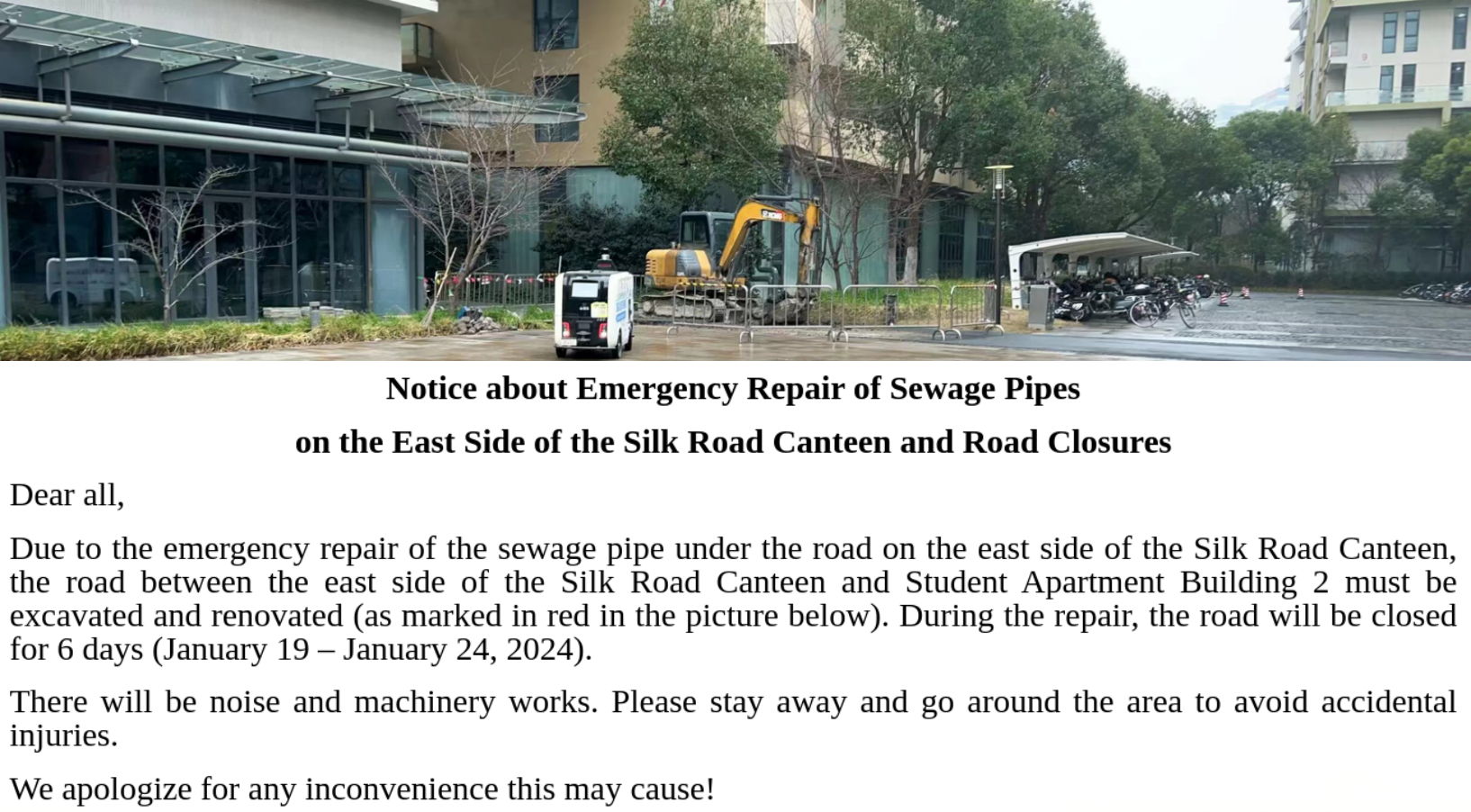}
	\caption{ The figure above depicts a real-life situation encountered by a 3rd-party delivery robot on our University campus, where it is blocked by an intersection closure. Below the notification sent by Office of General Services announcing this closure is shown.}	
	\label{fig:email}
	\vspace{-6mm}
\end{figure}

\vspace{-2mm}
Recent years have seen a growing interest in Large Language Models (LLMs) like ChatGPT\cite{OpenAI2022ChatGPT} and LLaMA\cite{touvron2023LLaMA}.
Real-life robots often face unpredictable situations, such as a campus delivery robot blocked by a closed intersection for pipe repair, depicted in Fig. \ref{fig:email}.
Despite the construction notice being posted publicly, the robot was unaware. Integrating general knowledge with real-time data from public notice boards through LLMs could significantly improve navigation and decision-making. For this integration to be effective, the robot's `brain' needs to grasp the map's hierarchy and topology.

\begin{figure}[t]
	\centering
	\includegraphics[width=0.45\textwidth]{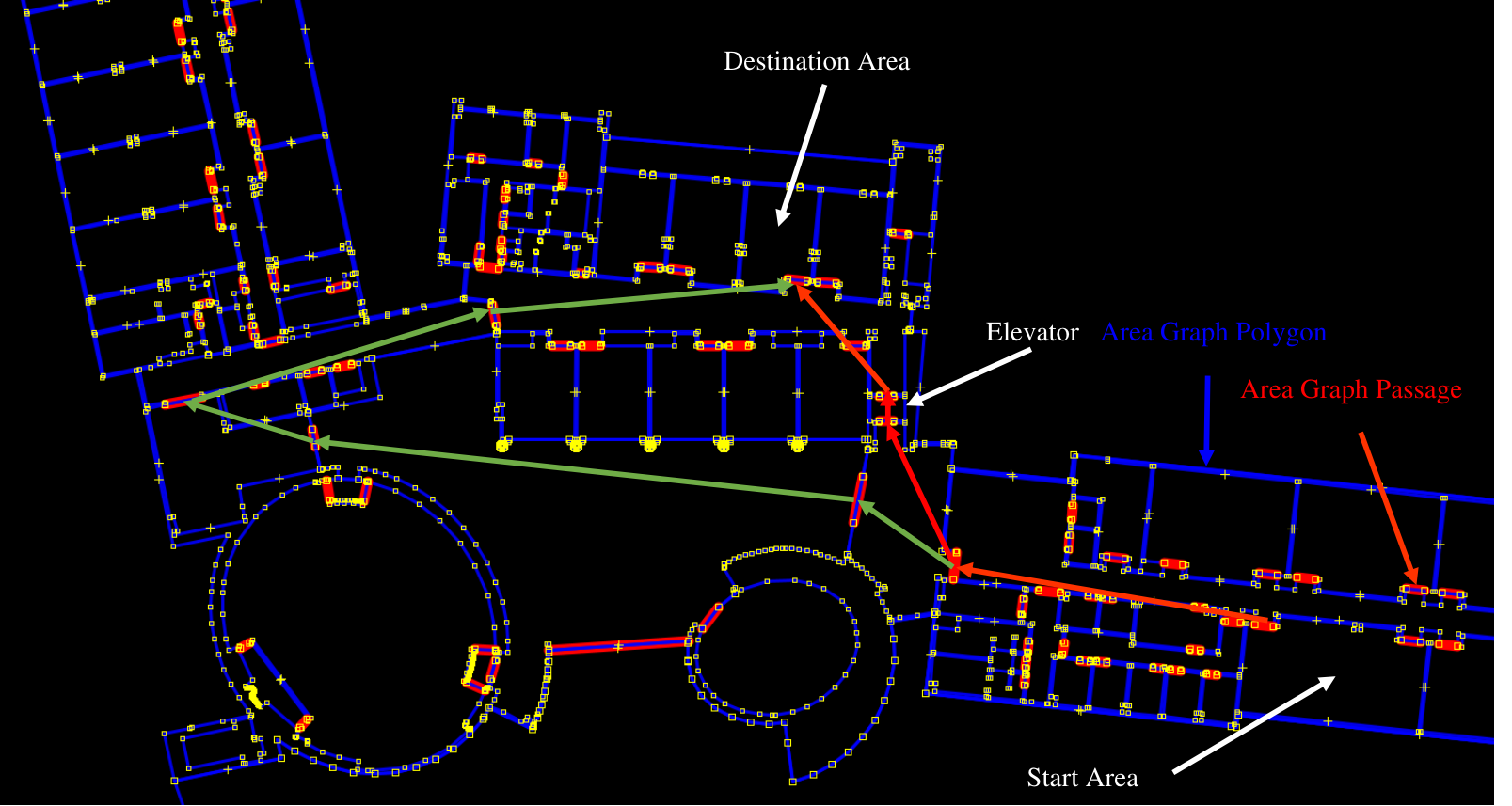}
	\caption{osmAG visualization in JOSM displays areas in blue polygons, passages in red lines, and nodes coordinates in yellow dots. ChatGPT-4 identifies the shortest path via a two-way elevator in red. After given an elevator maintenance notice, it recommends an alternate route in green. }
	\label{fig:area_graph}
	\vspace{-6mm}

\end{figure}

In mobile robotics, maps are essential foundational knowledge. Common robotic map formats include 2D occupancy grids, 3D point clouds, and visual approaches like bag of words, but these are suboptimal for LLM integration due to LLMs' text-based processing and token limitations. 
	We propose using osmAG\cite{feng2023osmAG} (Area Graph\cite{hou2019area}\cite{hou2022area} in OpenStreetMap format) for path planning in future mobile LLM-robot systems due to its advantages:

\begin{enumerate}
	
	\item osmAG is stored in text format, making it naturally readable by LLMs. 
	\item osmAG only stores permanent structures, ensuring long-term stability.
	\item osmAG can be easily generated from 3D point clouds\cite{he2021hierarchical}, 2D occupancy grid maps\cite{hou2019area}\cite{hou2022area}, or CAD files\cite{feng2023osmAG}.
	\item Conventional robotic localization\cite{xie2023robust} and path planning\cite{feng2023osmAG} algorithms based on osmAG have been developed, making LLM behavior easy to monitor and verify, thereby enhancing safety.
\item osmAG is easily visualized through JOSM and ROS's rviz, enabling intuitive human interaction with the map.

\end{enumerate}
The graph nodes of osmAG represent physical areas like rooms, while edges, termed passages, are door line segments of area polygons connecting adjacent areas.
For brevity, the details of  osmAG  are omitted in this paper.
Fig. \ref{fig:area_graph} presents a JOSM (Java OpenStreetMap editor) visualization of a osmAG map, as discussed in Section \ref{sec:real_life_experiment}.
The objective of this paper is to show that osmAG can be utilized by LLMs for applications in map-related mobile robotics tasks such as path planning.

Research on LLMs in mathematics has shown their challenges with numbers\cite{lewkowycz2022solving}\cite{mccoy2023embers}. Considering that humans often omit metric details in path planning discussions, as noted in the notification in Fig. \ref{fig:email}, excluding these metrics from maps does not diminish the system's effectiveness. Therefore, this paper focuses only on the topological and hierarchical properties of osmAG, disregarding its metric information.


For proprietary LLMs like ChatGPT, we adjust prompts and osmAG variants to evaluate their comprehension using datasets. For open-source LLMs, we fine-tune  LLaMA2 models, achieving over 90\% success rate in map comprehension tasks. Section \ref{sec:real_life_experiment} highlights ChatGPT-4's ability to understand real-life scenarios, demonstrating the utility of LLMs with osmAG in robotics.
Our contributions are as follows:
\begin{itemize}
	\item [$\bullet$] We propose utilizing osmAG as the map representation for future mobile LLM-robot systems.
	\item [$\bullet$] We offer scripts to convert osmAG for LLMs and generate datasets for fine-tuning on topology and hierarchy understanding.
	\item [$\bullet$] We employ efficient fine-tuning to achieve better performance with our LLaMA2 model than ChatGPT-3.5 on our tasks.
	\item [$\bullet$] Our dataset, dataset generation scripts, and LLaMA2 adapters are made publicly available to encourage further research and collaboration in the field.
	
\end{itemize}

\section{Related Works}
\subsection{Integrated Robotics and Navigation with LLMs and Maps}
Research in robotics integrates natural language models like PaLM-E\cite{driess2023palm}, which uses real-world sensor data for better decision-making, and LLMs for tasks such as code generation\cite{vemprala2023chatgpt}, object rearrangement\cite{ding2023task}, and motion planning\cite{sharma2022correcting}. Visual-language navigation systems like \cite{wang2023lana} allow robots to follow human instructions using advanced map technologies such as VLMaps\cite{huang2023visual} that combine language models with 3D point clouds. \cite{xie2024intelligent} integrates external information into robot navigation by utilizing osmAG as map representation. Our focus is on developing map representations that are easily understandable by both LLMs and robots, prioritizing map data to enhance navigation and decision-making in robotics.
\vspace{-1mm}
\subsection{Scene Graph in Robotics}
\vspace{-2mm}
Armeni et al. \cite{armeni20193d} and Hughes et al. \cite{hughes2022hydra} introduced Scene Graphs as RGB-D camera-based 3D representations that organize environments into layered graphs with nodes for spatial concepts from geometry to high-level semantics, and edges depicting relationships. Subsequent applications by Chen et al. \cite{chen2024scene}, and others have advanced Scene Graphs for localization and planning. Unlike Scene Graphs, osmAG, derived from grid maps, LiDAR, or CAD data, is less prone to occlusion and remains reliable without frequent updates. It focuses on permanent structures and doesn't require the semantic or visual data that traditional robot navigation algorithms use to avoid obstacles.
\begin{figure}[t]
	\centering
	\includegraphics[width=0.49\textwidth]{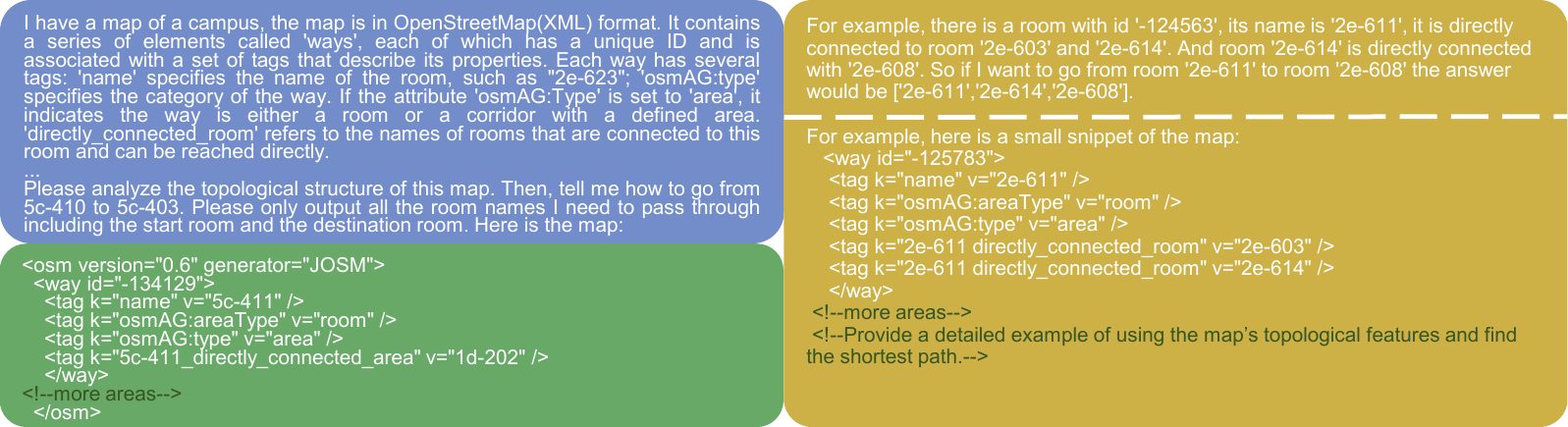}
	\caption{Our prompts include two main elements: a task description in a blue box and the osmAG map in a green box. We offer three prompt levels: Level 1 with just the description and map, Level 2 with a simple example in the upper yellow box, and Level 3 with a detailed example in the lower yellow box.}
	\label{fig:prompt_text}
	\vspace{-6mm}
\end{figure}

\section{Approach}

This paper explores how osmAG can aid LLMs in robotic tasks like path planning. For proprietary LLMs such as ChatGPT, in Sections \ref{subsec:PromptEngineering} and \ref{subsec:AreaGraphVariant}, we explore the levels of task description in prompt and the variants of osmAG, respectively. 
For open-source models like LLaMA2, initial results show a low success rate of approximately 0.1 (see Table \ref{table:fine_tune}). This indicates that merely combining prompt engineering with appropriate osmAG representation falls short of our objectives, leading us to also incorporate fine-tuning LLMs using our datasets.
Detailed in Section \ref{subsec:Dataset} is our methodology for dataset creation, and Section \ref{subsec:fine_tune} outlines the fine-tuning process using LoRA (Low Rank Adaptation)\cite{hu2021lora}. 
\vspace{-1mm}
\subsection{Prompt Engineering}
\label{subsec:PromptEngineering}
The prompt is structured into two parts: the first provides a complete task description, including the map format and task details, while the second part offers osmAG context. We explored three levels of detail in the task description (see Fig. \ref{fig:prompt_text}): all include basic osmAG and task explanations. Level 1 has no example, Level 2 includes a simple example, and Level 3 provides a detailed example with an illustrative map and detailed answer. Our aim is to enhance LLM performance by using examples to better clarify the task, leveraging in-context learning principles \cite{dong2022survey}.
\subsection{osmAG Variants}
\label{subsec:AreaGraphVariant}
\begin{figure}[t]
	\centering
	\includegraphics[width=0.48\textwidth]{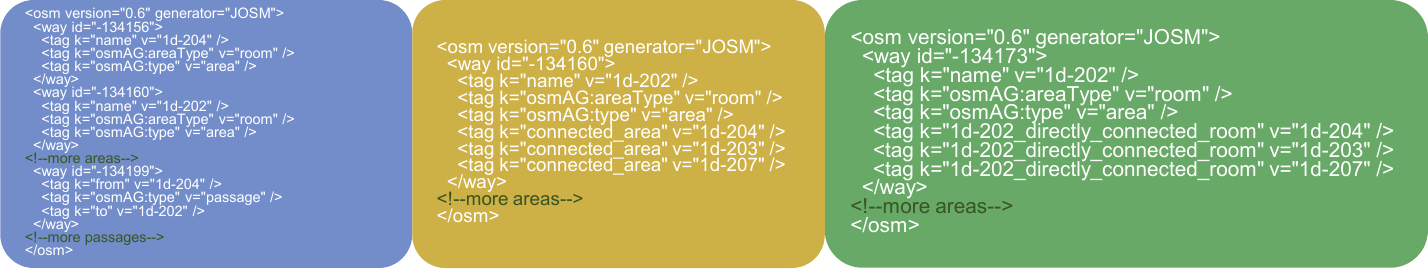}
	\caption{The osmAG map representation and two of its variants. The osmAG enclosed in blue box represents the original format that uses `passage' to describe connections between areas. In the yellow box we illustrate Variant 1 of osmAG, which introduces a tag with a key set to ``connected\_area" and a value corresponding to the area connected via the passage. Variant 2 of osmAG displayed in the green box, modifies Variant 1 by replacing ``connected\_area" with ``current area name\_directly\_connected\_room". }
	\label{fig:ag_varient}
	\vspace{-6mm}
\end{figure}

As shown in Fig. \ref{fig:ag_varient} and referenced in \cite{hou2019area}, the original Area Graph uses `passage' to connect different areas, with `passage' and `area' under separate `way' tags in the osmAG XML, complicating connection understanding for LLMs. We developed a script to create osmAG variants that integrate connection details directly within the `area' tags. Variant 1 adds a `connected\_area' key directly to the `area', specifying its connected area. Variant 2 further specifies the current area's name and its direct connections. 
\vspace{-1mm}

\subsection{Datasets }
\label{subsec:Dataset}
In order to test and fine-tune the ability of LLMs to understand the topology and hierarchy of osmAG, we need to construct specific datasets with language instructions, osmAG, and ground truth. 
We generate topological and hierarchical datasets specifically to enhance and evaluate the LLMs' capabilities in these areas.
Additionally, a general knowledge dataset is created to evaluate whether the model retains its general knowledge capabilities after fine-tuning.

Due to the limitation of token size (LLaMA2 supports up to 4096 tokens)
and our decision to omit the metric information of osmAG, the specific shapes of areas become irrelevant. Instead, only the information regarding connections and hierarchy remains pertinent. Consequently, employing hand-drawn layout templates is sufficient for evaluating the LLMs' proficiency in understanding topological and hierarchical relationships. 

\subsubsection{Topological Datasets}
\label{topo_dataset}
%

\begin{figure}[b]
	\vspace{-4mm}
	\centering
	\includegraphics[width=0.4\textwidth]{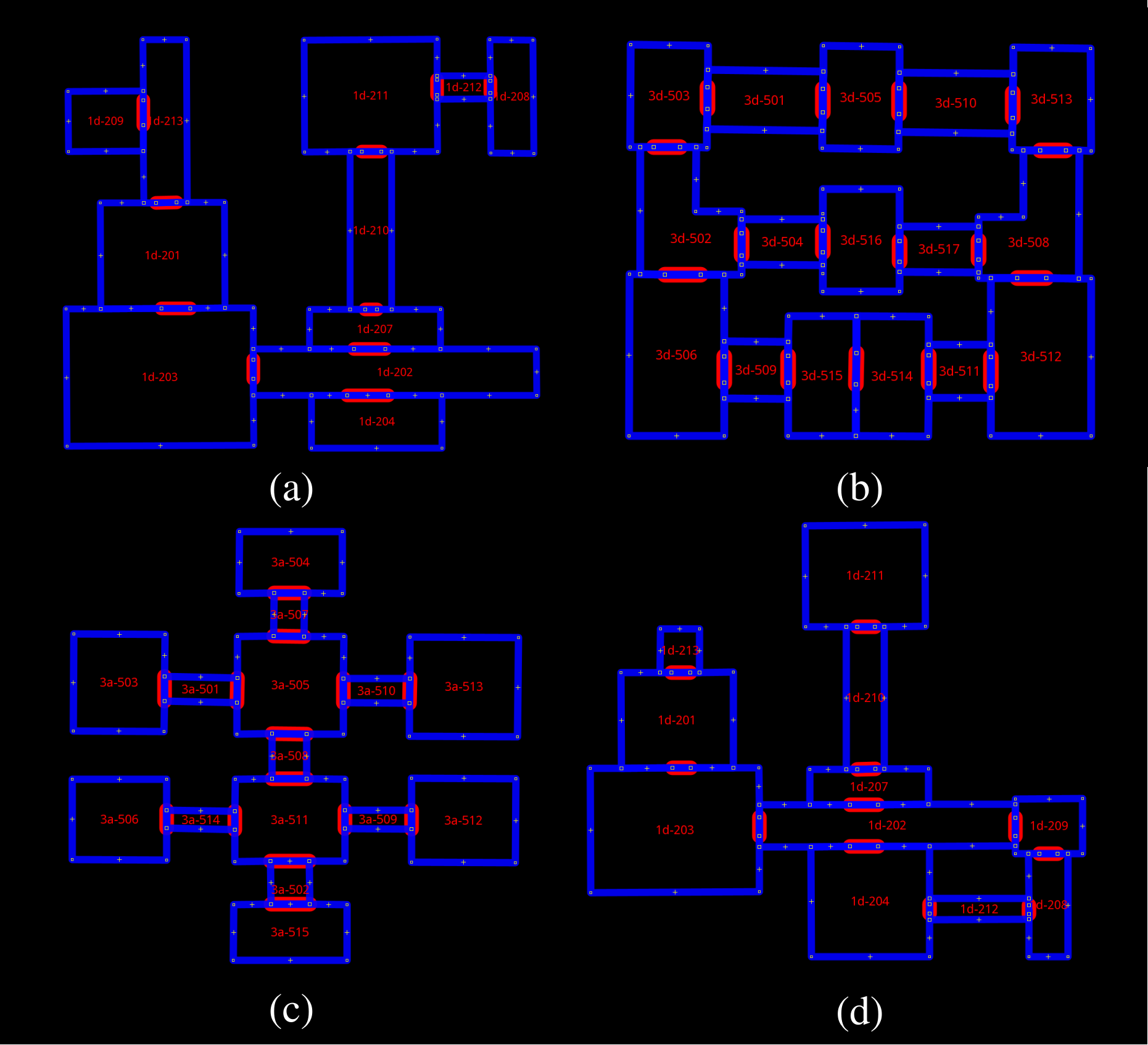}
	\caption{
		Hand-drawn map layout templates created using JOSM, with areas depicted as blue polygons, passages represented by red lines, and room names in red. Templates (a), (b), and (c) were utilized to generate datasets for fine-tuning the LLaMA2 model. Template (d), on the other hand, was exclusively used in the test dataset to assess the LLM's capacity to adapt to unseen layouts.
	}
	\label{fig:ag_temp}
	\vspace{-3mm}
\end{figure}

Map templates shown in Fig. \ref{fig:ag_temp} are handcrafted using JOSM, with `area' and `passage' defined per osmAG standards in \cite{feng2023osmAG}. Labels like `1d-201' indicate a room's location by zone, floor, wing, and room number. We employ a script to randomize these attributes to create varied maps from a single template. Room numbers are shuffled to ensure paths rely on map information, not sequence.
For each map, we ask the LLM to find a path (sequence of room names) between each two rooms as one item of the dataset. Dataset 1, using the `normal' layout Template (a) from Fig. \ref{fig:ag_temp}, contains 440 entries. Dataset 2, mixing Templates (a), (b), and (c), includes 12,520 entries aimed for training, with 440 reserved for testing. Dataset 3, derived from an Template (d) that is not used in the training dataset, aim to evaluate the model's generalization ability.
The number of rooms in templates is limited by token limits, preventing larger designs. 
However, leaf rooms with a single door are unnecessary for path planning unless they are the terminal rooms, so they can be omitted beforehand. Thus, despite the template's limited number of rooms, it remains highly relevant for real-world, large environment applications.

Based on experiment in Sections \ref{sec:prompt_engineer_exp} and \ref{sec:ag_variant_ex}, we utilize task description Level 3 and osmAG Variant 2 in prompts. 
Our datasets' ground truth is sequential room numbers from standard path planning algorithms. For circular paths with equal-length alternatives, both are included.

\subsubsection{Hierarchical Datasets}

\begin{figure}[t]
	\centering
	\includegraphics[width=0.45\textwidth]{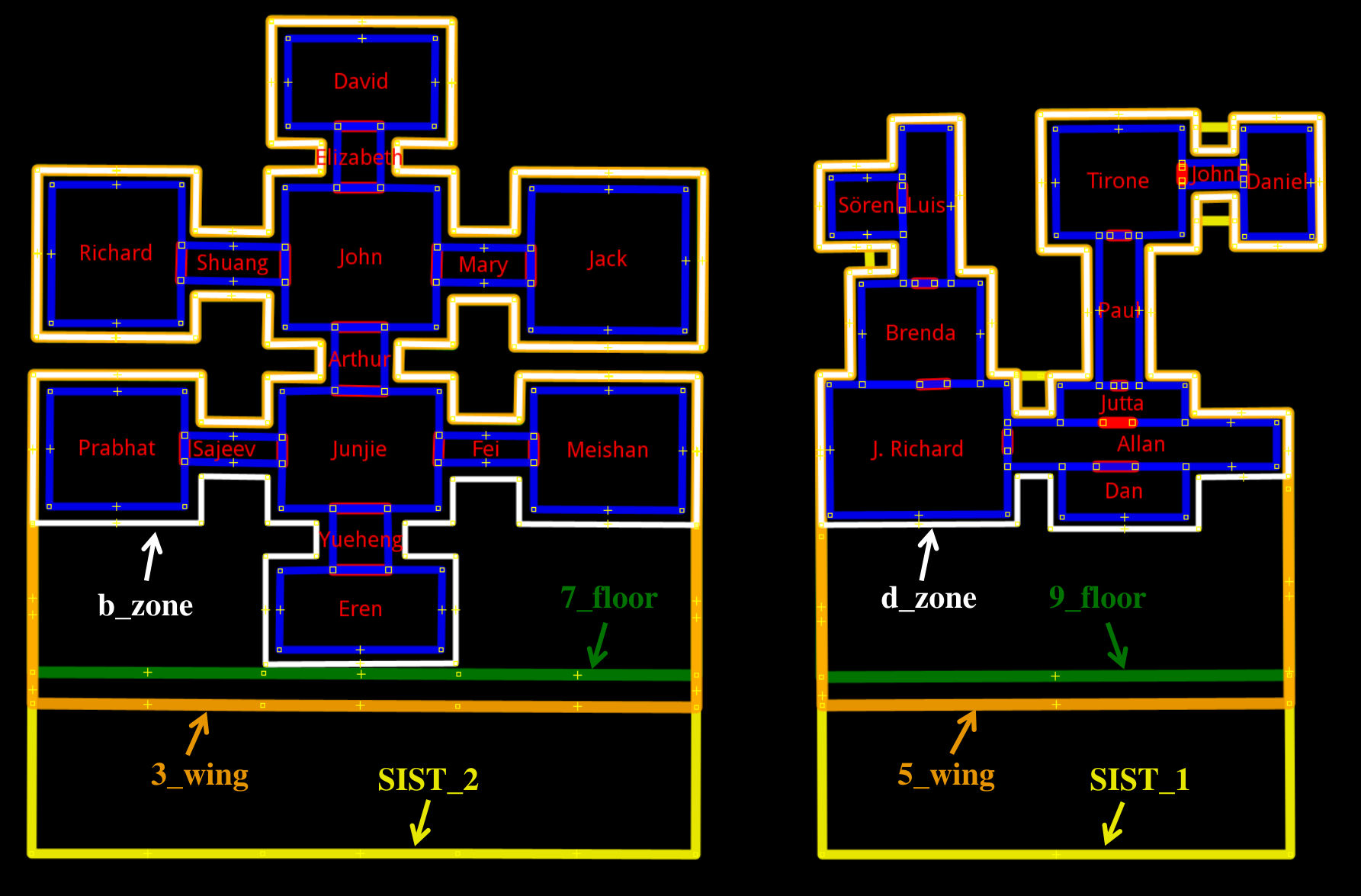}
	\caption{This image displays two osmAG maps from Dataset 2, each randomly assigned a `owner' tag per room (first names shown for clarity). The maps are structured by zone, floor, and wing, with buildings `SIST\_1' or `SIST\_2' to establish a hierarchical tree using `parent' tags. Details for the left map are: SIST\_2, 3\_wing, 7\_floor, b\_zone.}
	\label{fig:hier_map}
	\vspace{-6mm}
\end{figure}

As shown in Fig. \ref{fig:hier_map}, we randomly selected two osmAG from Dataset 2, assigning each room a random `owner' tag. We structured each map with `parent' tags linking rooms to zones, floors, and wings, and assigned buildings `SIST\_1' or `SIST\_2' to establish a hierarchy. This setup supports queries asking the LLM to locate individuals by directing towards the correct building. We created 1056 such queries for training Dataset 4. An example of this map visualized by JOSM is in Fig. \ref{fig:hier_map}.
The LLM must navigate hierarchy tree to correctly identify the target building.

\subsubsection{General Knowledge Dataset}

LoRA fine-tuning is not completely immune to catastrophic forgetting \cite{zhai2023investigating}. To verify that general knowledge capabilities remain intact, we created a small Dataset 5 with 20 general questions, like ``Who wrote `Hamlet'?", to test the model after fine-tuning.
%

\subsection{Fine-tuning}
\label{subsec:fine_tune}
%

\begin{figure}[t]
	\centering
	\includegraphics[width=0.40\textwidth]{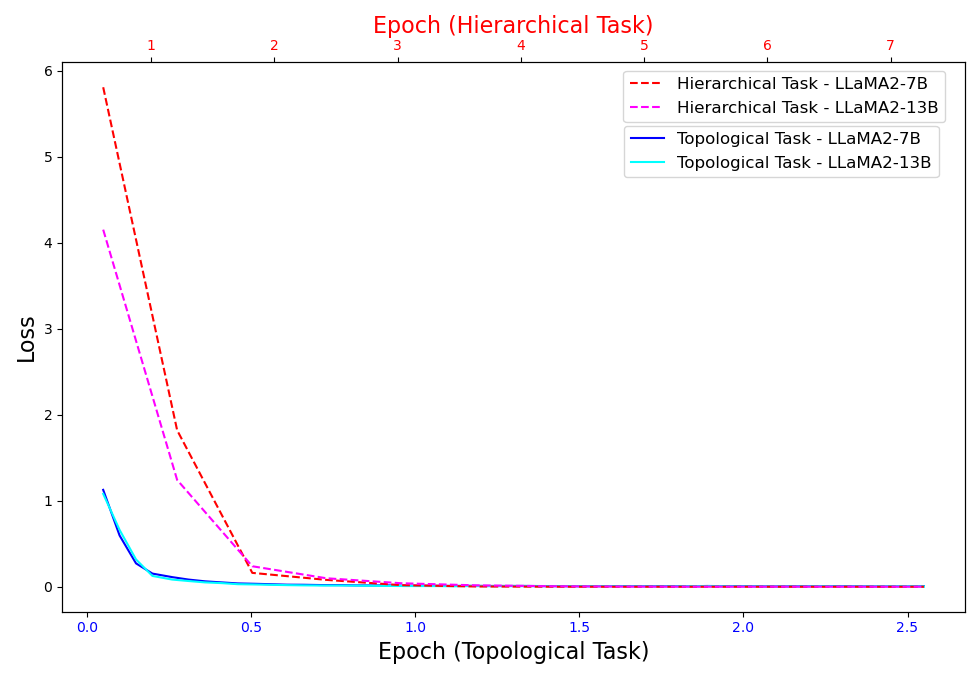}
	\caption{
		Fine-tuning LLaMA2-7B and LLaMA2-13B on Topological and Hierarchical Tasks 
	}
	\label{fig:loss}
	\vspace{-6mm}
\end{figure}

We utilize Meta's LLaMA2 model\cite{touvron2023LLaMA}, a decoder-only transformer, as our base model. The models are fine-tuned using datasets from Section \ref{subsec:Dataset} with the LoRA (Low Rank Adaptation)\cite{hu2021lora} method, which updates additive low-rank matrices in neural layers while keeping the original weights frozen, reducing the number of trainable parameters and allowing efficient use of computational resources. Fine-tuning targets topological tasks with Dataset 2 and hierarchical tasks with Dataset 4. 
The LoRA hyperparameters are as follows: the rank is set to 8, and target modules of LoRA are set to ``q\_proj" and ``v\_proj". These two modules are the query and value matrices in the self-attention mechanism of the transformer architecture \cite{hu2021lora,vaswani2017attention}. The learning rate is set to 5e-5 and we opted for a cosine learning rate scheduler.
The fine-tuning process is illustrated in Figure \ref{fig:loss}. The LLaMA2-7B and LLaMA2-13B models were fine-tuned for topological and hierarchical tasks using 4$\times$NVIDIA A40 and 4$\times$A100 GPUs, respectively. Topological tasks took 6.6 hours for LLaMA2-7B and 5.5 hours for LLaMA2-13B, both running 2.5 epochs with a dataset of 12,520 entries. Hierarchical tasks required 2 hours for LLaMA2-7B and 1.5 hours for LLaMA2-13B, each undergoing 7.76 epochs with 1,056 entries.
\vspace{-2mm}
\section{Experiments}
	\vspace{-2mm}
Since LLMs can only process textual input, comparing osmAG with occupancy grid maps or point clouds is impractical. Regarding the 3D Scene Graph, it primarily focuses on semantic representation within a 3D scene.
We assert that semantic information is unnecessary for mobile robot path planning between specified start and end rooms, and therefore, do not include comparisons with Scene Graph.

For topological tests, matching ground truth is considered successful, while correctly identifying buildings signifies success in hierarchical tests. Model performances are measured by success rates on the test dataset. Sections \ref{sec:prompt_engineer_exp} and \ref{sec:ag_variant_ex} detail experiments on ChatGPT-3.5 and ChatGPT-4 regarding prompt engineering and osmAG variants. Sections \ref{sec:without_ft_ex} and \ref{subsec:fine-tune_exp} compare pre and post-fine-tuning performance of LLaMA2 models. Section \ref{sec:real_life_experiment} and Fig. \ref{fig:area_graph} demonstrate ChatGPT-4 using real osmAG for live path planning and adjustments.
\begin{table}[t]
	
	\vspace{-0mm}
	\caption{{Comparison ChatGPT-3.5\&4's Success Rate on Different Prompt Levels and osmAG Variants }}
	\label{table:ag_varient}		
	\centering
	\setlength{\tabcolsep}{1.6mm}{
		\scalebox{0.79}{
			\begin{tabular}{ccccccc}
				\toprule
				\multirow{2}{1.4cm}{\shortstack{ Task \\Description \\   Level}}
				& \multicolumn{2}{c}{\shortstack{Original \\osmAG}}  &\multicolumn{2}{c}{\shortstack{osmAG \\Variant 1}}
				&\multicolumn{2}{c}{\shortstack{osmAG \\Variant 2}} \\ \cmidrule{2-7} 
				&\shortstack{Chat\\GPT3.5}&\shortstack{Chat\\GPT4}&\shortstack{Chat\\GPT3.5}&\shortstack{Chat\\GPT4}&\shortstack{Chat\\GPT3.5}&\shortstack{Chat\\GPT4}\\
				\midrule
				Level 1 &\bf{0.54}&0.85& 0.50&0.95&0.69&0.95  \\
				Level 2 &0.42&\bf{0.87}& 0.45& \bf{0.97}&\bf{0.70}&0.95\\
				Level 3& 0.49& \bf{0.87}& \bf{0.57}& 0.96&0.69&\bf{0.96}  \\

				\bottomrule
			\end{tabular}
	}}
	\vspace{-8mm}
\end{table}

\vspace{-1mm}
\begin{table*}[h!]
	\vspace{-6mm}
	\caption{{Comparison of Success Rates of LLMs on Topological (T), Hierarchical (H) and General (G) Tasks}}
	\label{table:fine_tune}		
	\centering
	\scalebox{0.89}{
		\begin{tabular}{ccccccc}
			\toprule
			&LLaMA2-7B&LLaMA2-13B&\shortstack{Fine-tuned LLaMA2-7B \\(with unseen prompt)}&\shortstack{Fine-tuned LLaMA2-13B \\(with unseen prompt)}&ChatGPT-3.5&ChatGPT-4.0\\
			\midrule
			$Dataset\ 1$ (T) &0.10&0.12& 0.99\ (0.78) &0.98\ (0.91)&0.54 &0.99 \\
			$Dataset\ 2$ (T) &0.05&0.066& 0.94\ (0.60)& 0.95\ (0.94)&0.50&0.89\\
			$Dataset\ 3$ (T) &0.11&0.14& 0.89\ (0.75)& 0.97\ (0.92)&0.53&0.96\\
			$Dataset\ 4$ (H) &0.19&0.55&1.0\ (0.98) &0.99\ (0.98) &0.66&0.99  \\
			$Dataset\ 5$ (G) &0.95&0.95& 0.95& 0.95&1&1\\
			
			\bottomrule
		\end{tabular}
	}
	\vspace{-6mm}
\end{table*}

However, it is important to note that, despite ChatGPT-4's high success rate, all its responses are verbose. This verbosity persists even when specifying concise outputs in the prompt, like 'only output the room numbers'. While this chattiness may be acceptable in human interactions, it poses challenges for traditional robotic applications to utilize the LLMs' responses. Nevertheless, we counted those answers as correct, if the room numbers matched the ground truth.


\subsection{Prompt Engineering Experiment}
\vspace{-1mm}
\label{sec:prompt_engineer_exp}

As discussed in Section \ref{subsec:PromptEngineering}, we tested ChatGPT-3.5 (gpt-3.5-turbo-0125) and ChatGPT-4 (gpt-4-0125-preview)'s understanding of osmAG path planning using three prompt levels. The results, shown in Table \ref{table:ag_varient}, reveal that including an example in the prompt does not always enhance performance. However, we opted for Level 3 prompts in our training dataset for topological tasks due to its marginally better results.
\vspace{-1mm}

\subsection{osmAG Variant Experiment}
\vspace{-1mm}
\label{sec:ag_variant_ex}

The osmAG variants for this experiment are detailed in Section \ref{subsec:AreaGraphVariant}. We compared original osmAG and two of its variants with different prompt levels on ChatGPT-3.5 and ChatGPT-4. According to the results summarized in Table \ref{table:ag_varient}, osmAG Variant 2 outperforms the others on both models. Therefore, we have selected this variant as the preferred map representation for LLMs and used it in our fine-tuning dataset.

\subsection{Topological \& Hierarchical Understanding Experiment Without Fine-tuning}
\label{sec:without_ft_ex}
We tested the LLaMA2-7B and LLaMA2-13B models on Datasets 1-5 to evaluate their map understanding. For comparison, we also used ChatGPT-3.5 and ChatGPT-4's APIs on these datasets. The results, shown in Table \ref{table:fine_tune}, reveal a 0.1 success rate for the LLaMA2 models in topological tasks without fine-tuning, which is impractical. ChatGPT-3.5 achieved about a 0.5 success rate, still insufficient for real-world use.
In hierarchical tasks, LLaMA2-13B achieved a 0.55 success rate, better than LLaMA2-7B's 0.19, but not yet practical. ChatGPT-3.5 reached a 0.66 success rate, still below practical deployment standards. ChatGPT-4 showed high success rates in both tasks.
\vspace{-2mm}

\subsection{Fine-tuning Experiment}

\label{subsec:fine-tune_exp}
After fine-tuning the LLaMA2 models as outlined in Section \ref{subsec:fine_tune}, we assessed their performance on Datasets 1-5, observing significant improvements detailed in Table \ref{table:fine_tune}.

For topological tasks, the fine-tuned LLaMA2-7B and LLaMA2-13B models exceeded ChatGPT-3.5's performance, with LLaMA2-13B also outperforming ChatGPT-4 on Dataset 2. LLaMA2-7B achieved over a 0.9 success rate on Datasets 1-2 with templates used in fine-tuning but dropped to 0.89 on new layouts. LLaMA2-13B showed stronger generalization with a 0.97 success rate on new layouts.

During testing, besides the fixed prompt from training, we also use random prompts to evaluate generalization. The results, shown in Table \ref{table:fine_tune}, reveal a performance drop for LLaMA2-7B under varied prompts, while LLaMA2-13B maintained high performance, suggesting its suitability for unpredictable interactions. Conversely, LLaMA2-7B is more suited for scenarios with consistent prompts, particularly when computational resources are limited.

In hierarchical tasks, after fine-tuning, both the LLaMA2-7B and LLaMA2-13B models achieve a success rate of 1, both models also generalize well on unseen prompts, which makes them totally practical.

\subsection{Real-life Experiment}
\label{sec:real_life_experiment}
Here we perform an experiment to emulate the real-world situation of a robot blocked by a construction site, as show in Fig. \ref{fig:email}. As depicted in Fig. \ref{fig:area_graph}, we send a request to the ChatGPT-4 API, providing a osmAG map with Prompt Level 3 to query a path planning from the start room to the destination room. The osmAG is converted into Variant 2, and any leaf areas with single door are removed to conserve tokens. ChatGPT-4 then returns a path, highlighted with red lines in the image, via a two-door elevator (identified by a semantic tag indicating it is an elevator). Upon introducing an email regarding elevator maintenance, ChatGPT-4 adjusts the path, adding a detour to bypass the unavailable elevator.

\section{Conclusion and Discussion}

In the rapidly evolving field of AI, LLMs are increasingly used to enhance robotic intelligence, though integrating them remains a key research area. This paper introduces osmAG, a map representation suited for LLM-robot systems, interpretable by LLMs, compatible with robotic algorithms, and understandable by humans.
For proprietary models like ChatGPT, we provide datasets to evaluate the model's comprehension of osmAG, along with osmAG variants to improve performance. For open-source models such as LLaMA2, we supply datasets, dataset generation methods, and fine-tuned adapters for comprehensive testing.
We recognize that in real robotics, path length is crucial, but current token limitations and the LLM’s difficulty with math mean we cannot ensure optimal paths. However, we are exploring closer integration of LLMs with traditional algorithms like A* \cite{feng2023osmAG} to address these issues.

Traditional robotics has been explored for decades, but integrating it with LLMs is a new frontier. osmAG aims to accelerate this integration, facilitating a map representation that aligns with LLMs, robotic systems, and human operators.
\vspace{-2.5mm}

%
%
\bibliographystyle{IEEEtran}
\bibliography{Bibliography}
\end{document}